\documentclass[runningheads]{llncs}

\let\llncssubparagraph\subparagraph
\let\subparagraph\paragraph
\usepackage[compact]{titlesec}
\let\subparagraph\llncssubparagraph
 
\usepackage{eccv}

\usepackage{floatrow}

\usepackage{epsfig}
\usepackage{graphicx}
\usepackage{amsmath}
\usepackage{amssymb}
\usepackage{multirow}
\usepackage{booktabs}
\usepackage{subcaption}
\usepackage{colortbl}

\definecolor{LightCyan}{rgb}{0.88,1,1}
\definecolor{Gray}{gray}{0.85}
\definecolor{LightGray}{gray}{0.5}
\definecolor{LightBlue}{rgb}{0.678,0.847,0.902} 
\usepackage[dvipsnames]{xcolor}

\usepackage[font=footnotesize,skip=0pt]{caption}

\usepackage{eccvabbrv}

\usepackage{graphicx}
\usepackage{booktabs}

\usepackage[accsupp]{axessibility}  

\usepackage{hyperref}

\usepackage{orcidlink}

\usepackage{titlesec}

\begin{document}

\title{Prompt Sliders for Fine-Grained Control, Editing and Erasing of Concepts in Diffusion Models} 

\titlerunning{P-S}

\author{Deepak Sridhar\inst{1}\orcidlink{0000-0003-4395-7366} \and
Nuno Vasconcelos\orcidlink{0000-0002-9024-4302}\inst{1}}

\authorrunning{D. Sridhar et al.}

\institute{University of California, San Diego, USA \\
\email{\{desridha,nvasconcelos\}@ucsd.edu}}

\maketitle

\begin{abstract}
  Diffusion models have recently surpassed GANs in image synthesis and editing, offering superior image quality and diversity. However, achieving precise control over attributes in generated images remains a challenge. Concept Sliders introduced a method for fine-grained image control and editing by learning concepts (attributes/objects). However, this approach adds parameters and increases inference time due to the loading and unloading of Low-Rank Adapters (LoRAs) used for learning concepts. These adapters are model-specific and require retraining for different architectures, such as Stable Diffusion (SD) v1.5 and SD-XL. In this paper, we propose a straightforward textual inversion method to learn concepts through text embeddings, which are generalizable across models that share the same text encoder, including different versions of the SD model. We refer to our method as \textit{Prompt Sliders}. Besides learning new concepts, we also show that prompt sliders can be used to erase undesirable concepts such as artistic styles or mature content. Our method is 30\% faster than using LoRAs because it eliminates the need to load and unload adapters and introduces no additional parameters aside from the target concept text embedding. Each concept embedding only requires 3KB of storage compared to the 8922 KB or more required for each LoRA adapter making our approach more computationally efficient. \textbf{Project Page:} \href{https://deepaksridhar.github.io/promptsliders.github.io/}{PromptSliders}
\end{abstract}

\section{Introduction}
\label{sec:intro}

\begin{figure*}[!ht]\RawFloats
\centering
\includegraphics[keepaspectratio, width=\columnwidth,trim=100 100 100 40, clip]{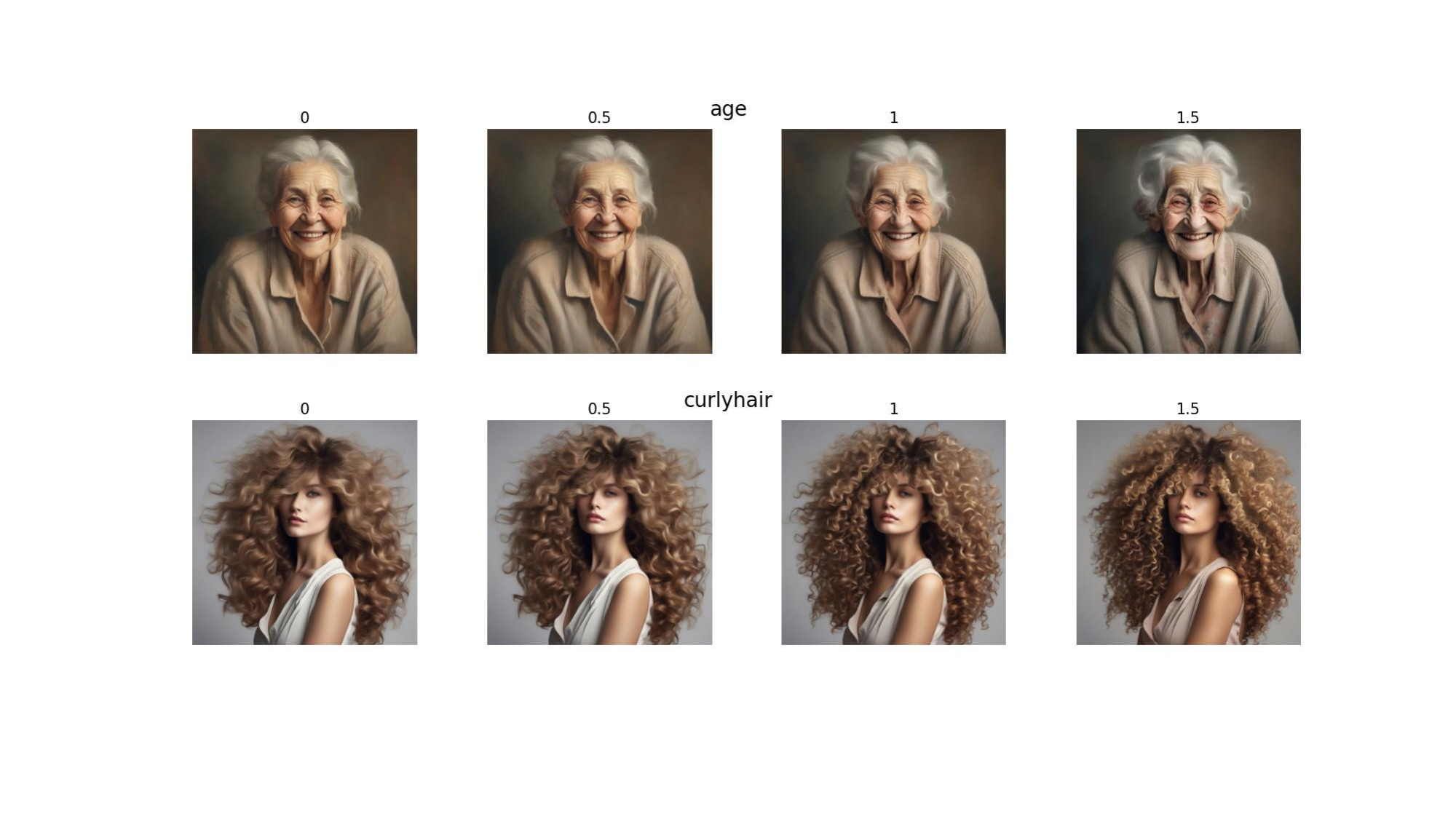}
\includegraphics[keepaspectratio, width=\columnwidth,trim=100 100 100 40, clip]{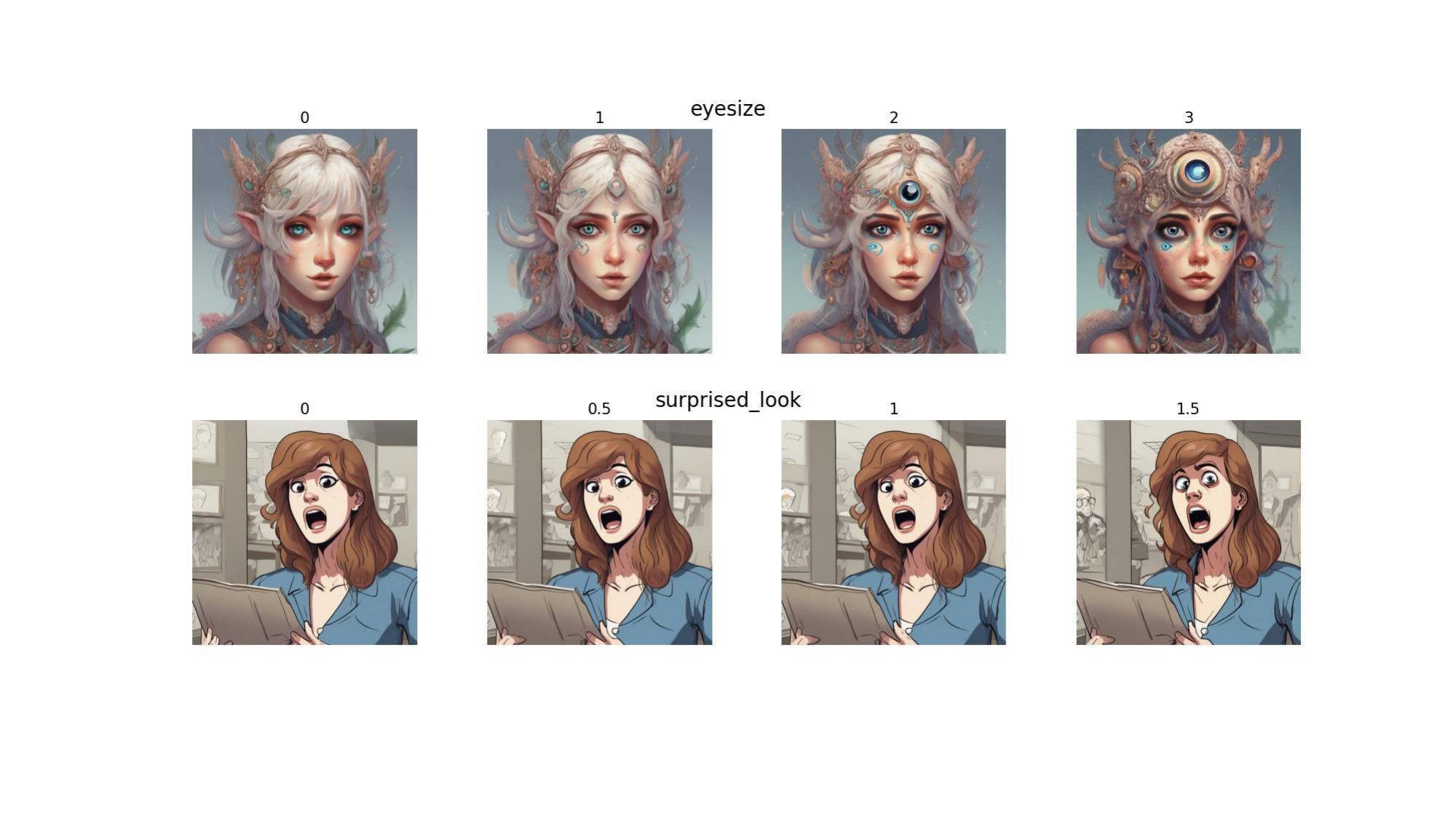}
\caption{Prompt Sliders for fine-grained control of attributes with textual inversion. Each row in the figure shows a corresponding concept depicted on top of the image along with its control strength $\alpha$ that enhances the target concept as the guidance strength increases. The prompts for the images from top to bottom are listed as follows (in order) - \textit{"A portrait of an woman with a warm smile", "A woman with voluminous hair cascading over her shoulders, posing for a fashion shoot", "A fantasy character", "A person caught off guard by unexpected news"}.
}
\label{fig:teaser}
\end{figure*}

Diffusion models (DMs) \cite{nethermo-sohl-dickstein15,ddpm_neurips_20, dhariwal2021diffusion, rombach2022high} have recently shown great promise for text-to-image (T2I) synthesis, where an image is generated in response to a text prompt. However, T2I synthesis offers limited control over image details. Even models trained at scale, such as Stable Diffusion (SD)~\cite{rombach2022high} or DALL-E 2~\cite{Ramesh2022HierarchicalTI}, have significant limitations, such as difficulty in modeling human fingers~\cite{fingers} or to control fine-grained attributes like age, smile etc. Recently, Concept Sliders \cite{gandikota2023sliders} demonstrated that learning a Low-Rank Adapter (LoRA) for a specific concept allows for much better control over the concept's strength in the generated image. However, LoRA adapters require millions of trainable parameters, necessitating higher GPU memory for training and increasing inference time when not merged into the base model. Because these adapters are designed to be plug-and-play, they must be loaded and unloaded for each concept as needed, adding 4 seconds to the inference time as reported in \cite{sdxllorainference}. This makes the method computationally expensive and significantly increases generation time—by 30\% for the SD-XL model \cite{podell2023sdxlimprovinglatentdiffusion}, for example. Furthermore, the trained weights are specific to the base model and cannot be used with other versions, such as Stable Diffusion 1.5 when the concepts are trained on SD-XL model.

\begin{figure*}[t]\RawFloats
\centering
\includegraphics[width=\columnwidth, trim=120 180 120 120, clip]{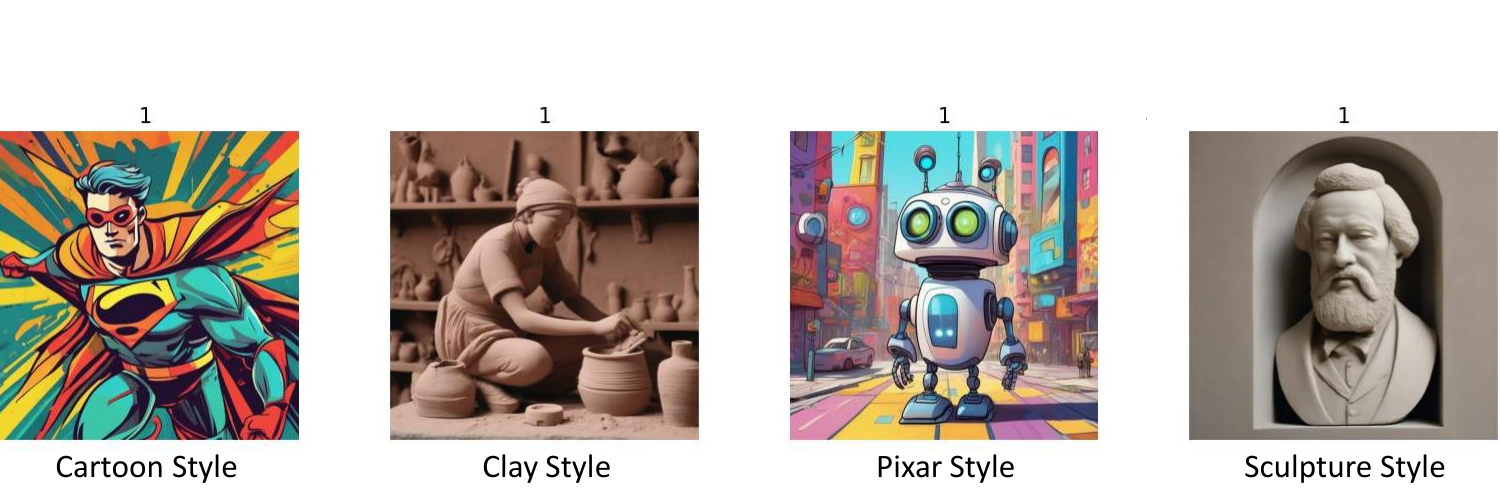}
\caption{\label{fig:style-teaser} Prompt sliders for encoding abstract concepts such as cartoon, clay, pixar and sculpture styles. The prompts used for generating the images are as follows (left to right in order) - \textit{"A superhero character in action, with bold lines and bright colors", "An artist working", "A funny and charming robot exploring a futuristic city", "A famous historical figure"}.}
\end{figure*}

In this paper, we address these issues by reformulating the concept slider as a textual inversion problem. We learn a text embedding that reflects the target concept and control the strength of the concept by the weight assigned to this text embedding. This is denoted as {\it prompt slider\/} and allows precise control over the generated concepts, as illustrated in Figure \ref{fig:teaser}, where learned concepts are enhanced as the guidance weight increases from left to right. Figure \ref{fig:style-teaser} shows different abstract concepts learned with this method. When compared to concept sliders, prompt sliders have a few advantages. First, the learned concept embeddings generalize across model architectures using the same text encoder. For example, the concept embedding learned for SD-XL is directly applicable to SD-1.4 or 1.5. Second, composing two or more concepts is straightforward, since each concepts can be controlled by its individual weight. In contrast, LoRA concept adapters are not trivial to merge \cite{yadav2023tiesmerging, yu2024languagedare} without losing the ability to independently control each concept. 

Overall, our contributions are as follows,

\begin{itemize}
    \item We propose a textual inversion method for learning concepts in diffusion models for fine-grained control, editing and erasing of concepts/attributes in the generated images.
    \item We show that the proposed method is flexible and generalizable across model architectures sharing the same text encoders.
    \item We demonstrate that combining multiple concept embeddings is straightforward with text prompts and allows for independent control of each concept, unlike combining multiple adapters.
\end{itemize}

\section{Related Work}
\textbf{Deep Image Generative Models} learn to synthesize images from noise. These models include generative adversarial networks
(GANs) \cite{goodfellow_2014_gan,brock2018large,karras2019style,Karras_2020_CVPR}, Variational Autoencoders (VAEs) \cite{kingma2013auto, vae-rezende14}, DMs \cite{ho2020denoising,nichol2021improved} and poisson flow generative methods \cite{xu2022poisson, pmlr-v202-xu23m}. 
 Latent DMs (LDMs) \cite{rombach2022high} implement DMs in the latent space learned by an autoencoder trained on a very large image dataset, to  reduce inference and training costs. 

 \paragraph{\bf Model Editing} Model Editing aims to finetune or edit a pre-trained foundation model to achieve higher control and remove harmful concepts from the base model. Erasing \cite{gandikota2023erasing, leco2023} developed low rank implementations for erasing concepts from diffusion models allowing the ability to adjust
the strength of erasure in the model. \cite{sdxltricks2023} implemented image
based control of concepts by merging two overfitted LoRAs
to capture an edit direction. Similarly, \cite{gandikota2024unified, kim2023safeselfdistillationinternetscaletexttoimage} proposed
closed-form formulation solutions for debiasing, redacting
or moderating concepts within the model’s cross-attention
weights. Concept Sliders \cite{gandikota2023sliders} learn a low-rank adapter which singles
out a semantic attribute and allows for continuous control
with respect to the attribute. They can be applied as a plug-and-play
module stacked across different attributes. Unlike these methods, our approach aims to modify the outputs of a generative model by manipulating the input textual space.

\paragraph{\bf Personalization} methods 
seek to learn a user-defined concept not commonly found in the training data for discriminative \cite{eccv2022_palavra_cohen} or generative \cite{nitzan2022mystyle} tasks. Recently, personalization of T2I models has gained significant attention due to the ease with which new concepts can be embedded into pre-trained diffusion models and generated for various contexts. Textual Inversion (TI) \cite{gal2022textual} represents the new concept as a learnable text token and optimizes the token embedding using a few example images (typically 3-5 per concept) to reconstruct the concept using the standard diffusion loss, while keeping the model weights unchanged. DreamBooth (DB) \cite{ruiz2022dreambooth} identifies a rare token from the existing vocabulary and fine-tunes the model weights to reconstruct the concept. DB adds a class-specific preservation loss to avoid language drift and maintain the diversity of the pretrained model. Custom Diffusion \cite{kumari2022customdiffusion} proposes a method for multi-concept composition by fine-tuning only a subset of the attention layers. LoRA \cite{hu2022lora, ryu2022} uses low rank adapters to finetune pretrained models on the target dataset. Perfusion \cite{tewel2023keylocked} avoids overfitting by introducing a dynamic rank-1 update that “locks” new concepts’ cross-attention keys to their superordinate category. SVDiff \cite{han2023svdiffcompactparameterspace} decomposes the weight kernels by singular-value decomposition resulting in an efficient personalization method in the parameter space. To avoid entanglement of styles for difficult compositions or similar category subjects, they propose a mixing and unmixing regularization. Break-a-Scene \cite{breakascene} extends this to multiple objects and operates on a single image containing multiple concepts.
Many fast personalization methods that have been introduced recently using dedicated encoders \cite{chen2023subjectdriven, gal2023encoderbaseddomaintuningfast, jia2023tamingencoderzerofinetuning, Shi_2024_CVPR, wei2023elite} can also handle a single image. P+ \cite{voynov2023P+} extends TI to utilize a richer inversion space. In this paper, we use textual inversion to learn a concept embedding and control its strength using a guidance weight.

\paragraph{\bf Image Editing} Recent techniques \cite{shi2023dragdiffusion, Kim_2022_CVPR, nichol2022glidephotorealisticimagegeneration, Geng23instructdiff} have introduced various approaches for single-image editing in T2I diffusion models. Many of these methods focus on manipulating cross-attentions between a source image and a target prompt \cite{hertz2022prompt, kawar2023imagic, zero-shot-im-im-parmar}. SDEedit \cite{meng2022sdedit} adds Gaussian noise to the conditional inputs used to guide the image structure and then runs the reverse SDE to synthesize images. Park et al. \cite{park2023understanding} identified local latent bases that facilitate semantic editing by navigating the latent space. By examining diffusion models through the lens of Riemannian geometry, their analysis also uncovered the evolving geometric structure over time across prompts, necessitating per-image latent basis optimization. Instruct-pix2pix \cite{brooks2022instructpix2pix} fine-tunes a diffusion model using editing instruction data, conditioning image generation on both an input image and edit instructions. While this allows for natural language-based image editing, it lacks fine-grained control over the strength of edits or visual concepts that are difficult to articulate textually. Unlike prior methods that are applied to a single image or a single model, our
method learns a semantic attribute for a given text encoder.
This allows the learned textual embeddings to be  generalizable across different models (sharing the same text encoder), without needing custom optimization for each model and image.

\section{Method} 

\subsection{Background on Diffusion Models}

Diffusion Models~\cite{nethermo-sohl-dickstein15,ddpm_neurips_20} are probabilistic models based on two Markov chains. In the forward direction,  white Gaussian noise is recursively added to image $x$, according to 
\begin{eqnarray}
\mathbf{z}_t = \sqrt{\alpha_t} \mathbf{z}_0 + \sqrt{1-\alpha_t} \epsilon_t,  \quad \epsilon_t \sim {\cal N}( \mathbf{0}, \mathbf{I}),
\label{eq:zt}
\end{eqnarray}
where $\mathbf{z}_0=x$, ${\cal N}$ is the normal distribution, $\mathbf{I}$ the identity matrix,  $\alpha_t = \prod_{k=1}^t (1-\beta_k)$, and $\beta_t$ a pre-specified variance.
In the reverse process, a neural network $\epsilon_\theta(\mathbf{z}_t, t)$ recurrently denoises $\mathbf{z}_t$ to recover $x$. 
This network is trained to predict noise $\epsilon_t$, by minimizing the risk defined by the loss 
\begin{eqnarray}
    \mathcal{L} = ||\epsilon_t - \epsilon_\theta(\mathbf{z}_t, t)||^2.
\end{eqnarray}
Samples are generated with $ \mathbf{z}_{t-1} = f( \mathbf{z}_{t}, \epsilon_\theta(\mathbf{z}_t, t))$ where
\begin{eqnarray}
     f( \mathbf{z}_{t}, \epsilon_\theta) = \frac{1}{\sqrt{\alpha_t}} \left(\mathbf{z}_t - 
    \frac{\beta_t}{\sqrt{1-\alpha_t}} \epsilon_\theta\right) + \sigma \xi,
    \label{eq:zt-1}
\end{eqnarray}
with $\xi \sim {\cal N}( \mathbf{0}, \mathbf{I}), \mathbf{z}_{T} \sim {\cal N}( \mathbf{0}, \mathbf{I})$. The network
$\epsilon_\theta(\mathbf{z}_t, t)$ is usually a U-Net~\cite{ronneberger2015u} with attention layers~\cite{vaswani2017attention}.

In Latent Diffusion Models (LDMs) \cite{rombach2022high}, an encoder $\mathcal{E}$ is trained to transform images \(x \in \mathcal{D}_x\) into a spatial latent representation \(z = \mathcal{E}(x)\), which is regularized either by a KL-divergence loss or by vector quantization \cite{VanDenOord2017, Agustsson2017}. The decoder \(\mathcal{D}\) is then trained to reconstruct images from these latent representations, ensuring that \(\mathcal{D}(\mathcal{E}(x)) \approx x\). The latent representation of the image is then used for training the diffusion models keeping the autoencoder frozen.



\begin{figure*}[t]\RawFloats
\centering
\includegraphics[keepaspectratio, width=0.8\columnwidth,trim=68 68 8 68, clip]{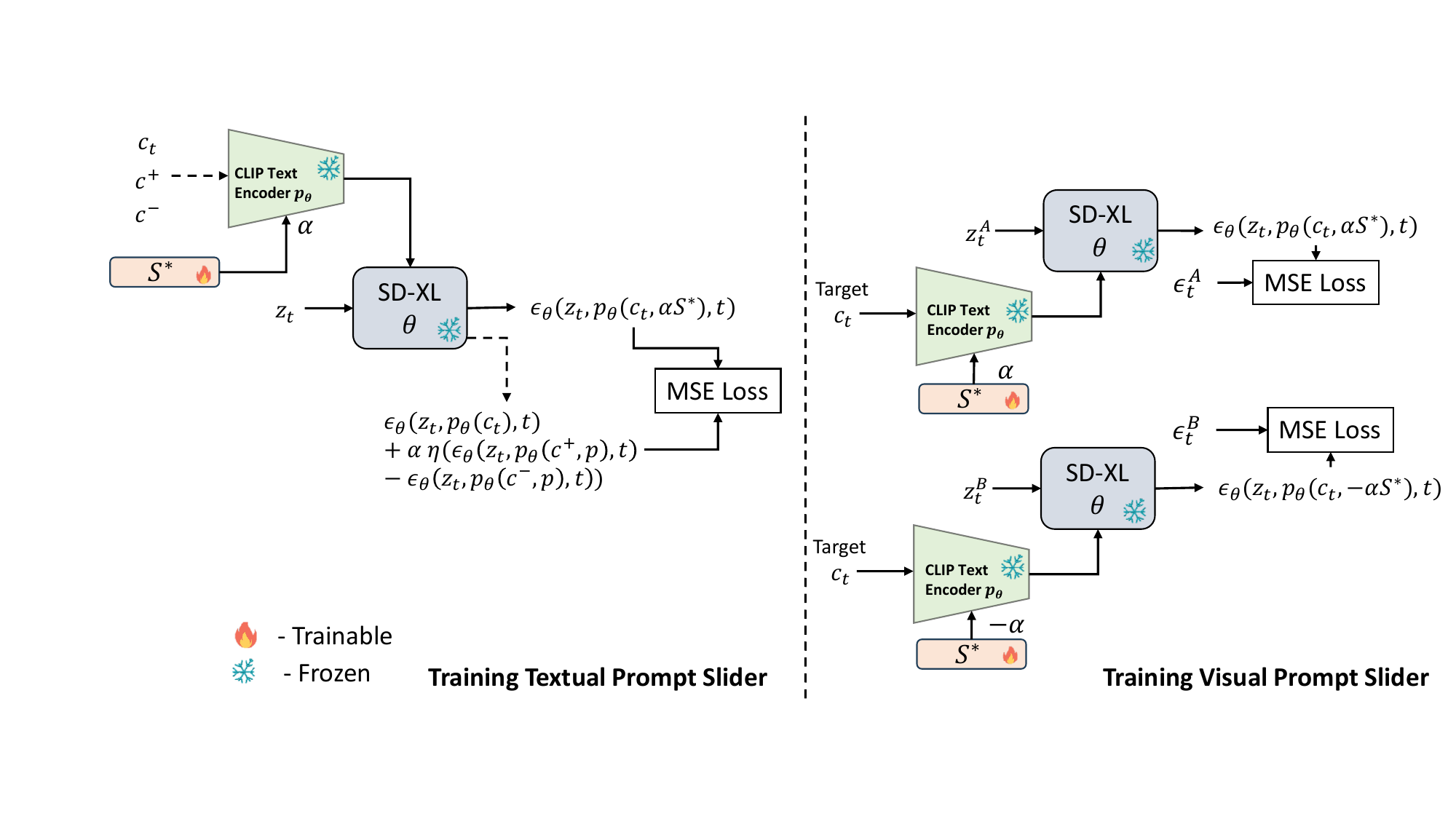}
\caption{\textbf{Left:} Training of textual Prompt Sliders. \textbf{Right:} Training of visual Prompt Sliders. 
}
\label{fig:arch}
\end{figure*}
\subsection{Textual Prompt Slider}
Concept Sliders are a technique for fine-tuning a diffusion model to achieve concept-specific image manipulation, using LoRA adapters. This method identifies low-rank parameter directions that can enhance or diminish the representation of particular attributes when conditioned on a designated concept. While LoRA adapters are small compared to the overall network size, they still contain a sizeable number of parameters. For example, 4.3 M for the ones used in~\cite{gandikota2023sliders}. This makes the loading and unloading of adapters costly, adding significant time to image synthesis. Textual Inversion is a more efficient adaptation technique, which learns new tokens in the text embedding space of a pre-trained diffusion model to represent a target concept. Let \( p_{\theta}\) be an embedding that maps text input \( y \) into a conditioning vector \( p_{\theta}(y) \). Textual inversion learns a prompt $S^*$ representative of the target concept through the optimization 
\begin{equation}\label{eq4}
S^* = \arg\min_S \mathbb{E}_{z \sim \mathcal{E}(x), y, \epsilon \sim \mathcal{N}(0,1), t} \left\| \epsilon_t - \epsilon_{\theta}(z_t, p_{\theta}(y, S), t) \right\|_2^2,
\end{equation}
where \(t\) is the time step, \(z_t\) is the latent noised to time \(t\), \(\epsilon_t\) is the noise sample, and \(\epsilon_{\theta}\) is the denoising network. Both \(p_{\theta}\) and \(\epsilon_{\theta}\) are fixed during the optimization.
We propose to use textual inversion for controlling concept/attribute strength.  This consists of embedding the target concepts/attributes in the text embedding space and controlling their strength by tuning the weight of the learned token embedding, using a technique similar to \cite{gandikota2023sliders}. 

Given a target concept $c_t$, we propose to learn the corresponding textual embedding $S^* \in \mathcal{R}^d$ ($d=768$ for CLIP text encoder) 
that encourages the distribution of \(c_t\) to exhibit more positive attributes \(c^+\) and fewer negative attributes \(c^-\). As illustrated in the left of Figure~\ref{fig:arch}, this is implemented by replacing $\epsilon_t$ by 
\begin{equation}\label{eq5}
\resizebox{\textwidth}{!}{$
\epsilon_t(\alpha) = \epsilon_{\theta}(z_t, p_{\theta}(c_t), t) +
\alpha \eta \sum_{p \in P}
(\epsilon_{\theta}(z_t, p_{\theta}(c^+, p), t) - \epsilon_{\theta}(z_t, p_{\theta}(c^-, p), t))
$}
\end{equation}
and $S$ by $\alpha S$ in~(\ref{eq4}), where $\eta$ is a guidance scale, $\alpha$ a scaling parameter, and $P$ a set of concepts that the attribute manipulation should preserve (for example,
race or gender). The positive \(c^+\), and negative \(c^-\) prompts are sampled from a template predefined for concept $c_t$. The scaling parameter $\alpha$ controls the strength of the edit, as shown in Figure \ref{fig:teaser}. During training, we randomly sample $\alpha$ from a given minimum and maximum scale. We set the minimum scale to be 0 and maximum scale to be 3 in all our experiments. During inference, increasing $\alpha$ makes the edit stronger as shown in Fig. \ref{fig:teaser}.

\subsection{Visual Prompt Slider}
For fine-grained visual concepts such as styles or fine-grained attributes like eyebrow that are difficult to describe by text, \cite{gandikota2023sliders} uses positive/negative ($z^A$/$z^B$) image pairs to define the positive and negative directions in the low rank adapters. Similarly, Prompt Sliders can also be trained for such concepts by minimizing the loss
\begin{equation}\label{eq6}
\left\|  \epsilon_t^A - \epsilon_{\theta}(z_{t}^{A}, p_{\theta}(c_t, \alpha S), t) \right\|^2 + \left\| \epsilon_t^B - \epsilon_{\theta}(z_{t}^{B}, p_{\theta}(c_t, -\alpha S), t) \right\|^2
\end{equation}
where a positive $\alpha$ is used to align the text embedding to the visual concept in image A and a negative $\alpha$ is used to align the text embedding to the negative image B. This is illustrated in the right of Figure~\ref{fig:arch}.


\subsection{Erasing Concepts}
Since the formulation of (\ref{eq5}) allows one to enhance a particular concept, inverting it allows one to erase the concept. Formally, this is defined as using a negative $\alpha$ in (\ref{eq5}),
\begin{equation}\label{eq7}
\resizebox{\textwidth}{!}{$
\epsilon(\alpha) = \epsilon_{\theta}(z_t, p_{\theta}(c_t), t) -
\alpha \eta \sum_{p \in P}
(\epsilon_{\theta}(z_t, p_{\theta}(c^+, p), t) - \epsilon_{\theta}(z_t, p_{\theta}(c^-, p), t))
$}
\end{equation}
Here, $c^+$ is the target concept to be erased while $c^-$ can be null or negative condition to the target. 
\section{Experimental Results}
 \paragraph{\bf Models and Evaluation:} We consider SD-XL, 1024x1024 model as the baseline for all experiments. We randomly sample $\alpha$ between 0 and 3 during training and use 50 DDIM inference steps with a classifier guidance scale of 7.5 for all the results presented. We evaluate the learned new token embeddings on a pretrained SD v1.4 and SD v1.5 models with 512 input resolution. We evaluate alignment of the generated image with the prompt using the CLIP score (CLIP-s) \cite{radford2021learning}.
 

\paragraph{\bf Implementation details}
We train all prompt sliders for 3000 iterations with a batch size of 1 on one NVIDIA-A40 GPU. Total training time is around 30 minutes per concept with bf16 precision. We use a constant learning rate of 5e-4 with AdamW optimizer with a weight decay of 0.01 and beta values (0.9, 0.999).

\begin{figure*}[t]\RawFloats
\centering
\includegraphics[width=\columnwidth, trim= 0 120 0 120, clip]{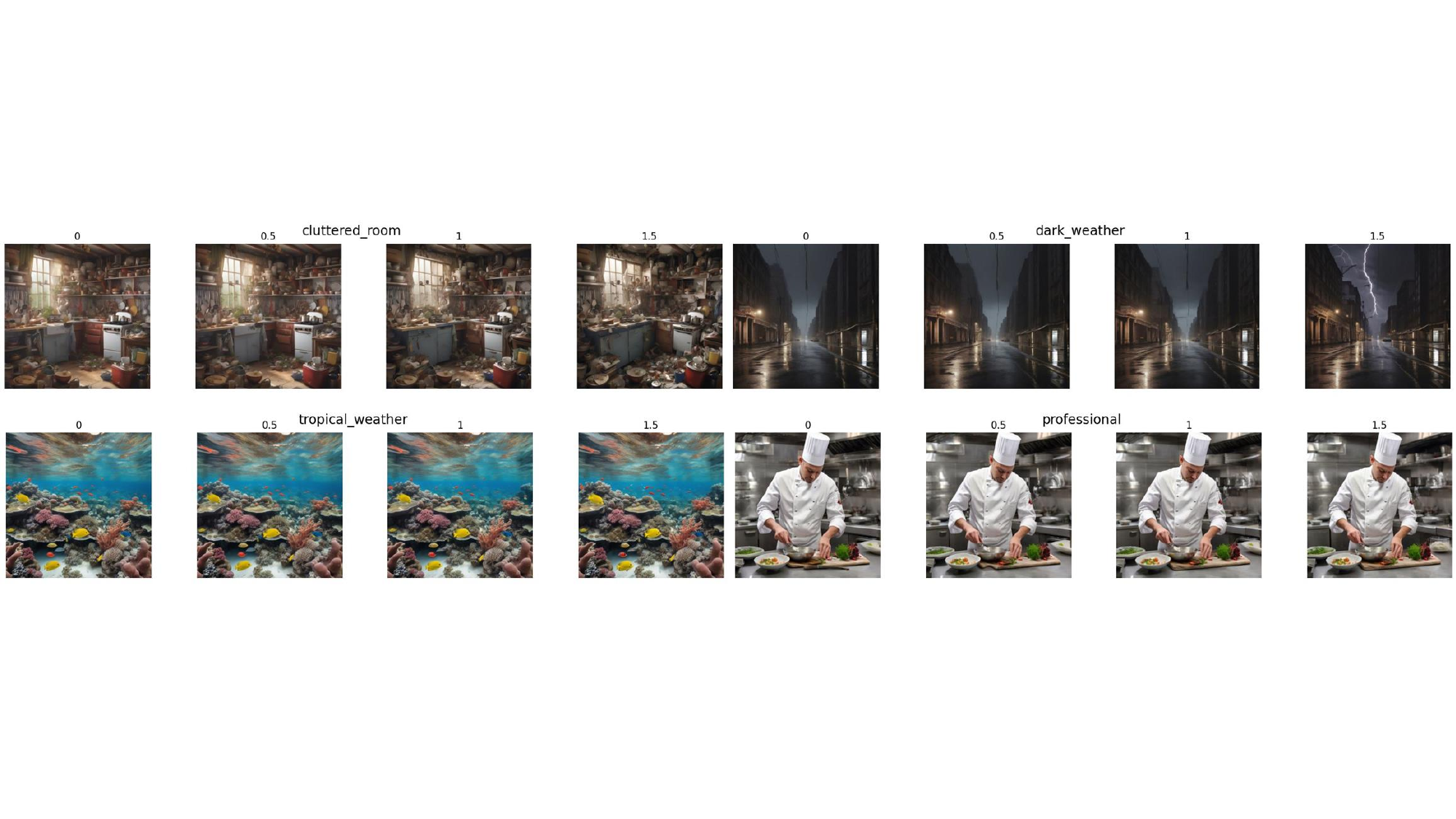}
\caption{\label{fig:qual} \textbf{More qualitative results of Prompt Sliders} depicting various concepts. The corresponding prompts for the images in the figure from left to right in the top row, and from left to right in the bottom row are as follows. \textit{"A kitchen", "A building in a city", "A vibrant coral reef teeming with marine life, seen through crystal-clear water", "A chef in a kitchen, skillfully preparing a gourmet dish"}.}
\end{figure*}

\subsection{Qualitative Results} 
Figure \ref{fig:qual} shows additional qualitative results of textual prompt sliders depicting various concepts such as cluttered room, dark weather, tropical weather etc. The bottom row of Fig. \ref{fig:qual} shows examples of static concepts that do not vary with the control strength $\alpha$.

\subsection{\bf Quantitative Results and Computation Complexity} 


Table \ref{tab:speed} shows the clip score and inference cost in terms of FLOPs and time taken for single image generation. CLIP score is evaluated on a set of 100 prompts with 5 prompts for each concept. It can be seen that Prompt Sliders notably improve the CLIP score compared to the baseline while maintaining the generation speed. They also add no inference cost to the base SD-XL model. This is unlike  Concept Sliders, which achieve a lower CLIP score and increase inference cost substantially (by 31\%). Additionally, while each concept slider requires about 8922 KB of storage, prompt sliders require only \textbf{3.1 KB (a 3000-fold decrease)}. This is because the concept embeddings only have 768 trainable parameters.

\begin{figure*}[t]\RawFloats
\centering
\captionof{table}{Comparison of CLIP-score and inference times for LoRA based sliders against the proposed prompt sliders.}

\begin{minipage}{\columnwidth}
\centering
\scriptsize
\setlength{\tabcolsep}{1pt}
\begin{tabular}{ l| l | l |l } 
\toprule
 SD-XL&\#P (M) & Time (s) $\downarrow$ & CLIP-s $\uparrow$\\
\midrule
Base \cite{podell2023sdxlimprovinglatentdiffusion} & 2567.5&  9 & 28.90\\
LoRA Slider \cite{gandikota2023sliders} & 2571.8 & 13 & 28.52\\
Prompt Slider (Ours)&     \textbf{2567.5}&  \textbf{9 (31\% $\downarrow$)} & \textbf{30.00}\\
\midrule
 SD-v1.5&\#P (M) & Time (s) $\downarrow$ & CLIP-s $\uparrow$\\
 \midrule
 Base \cite{rombach2022high} & 859&  4 & 21.36\\
Prompt Slider (Ours) & 859 & 4 & \textbf{23.93}\\
Prompt Slider (Ours) &     859&  4 & \textbf{23.97}\\
\multicolumn{1}{c}{\text{(Transferred from SD-XL)}} \\

\bottomrule
\end{tabular}
\label{tab:speed}
\end{minipage}
\end{figure*}

\begin{figure*}[t]\RawFloats
\centering
\includegraphics[width=\columnwidth, trim= 0 60 0 60, clip]{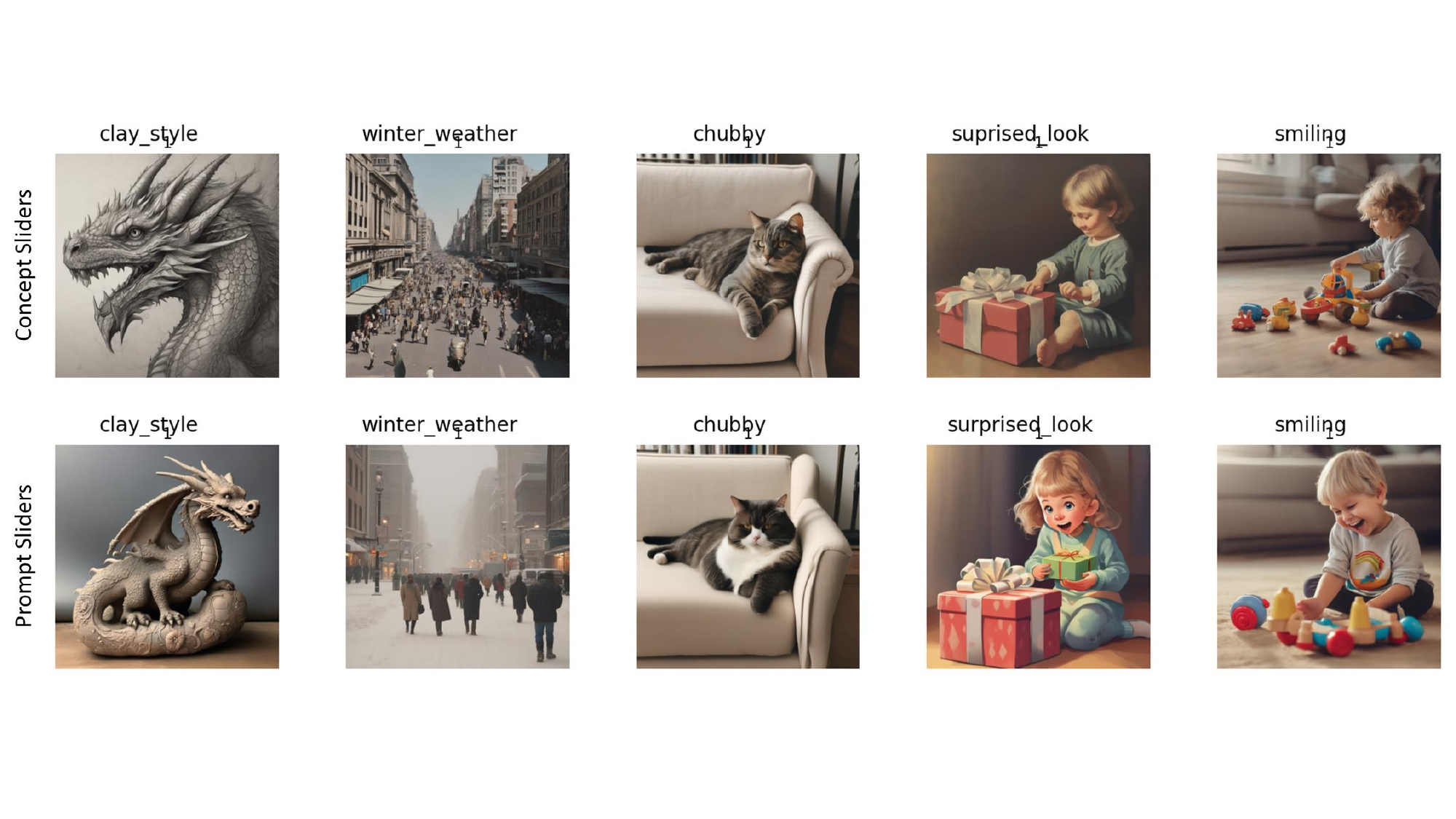}
\caption{\label{fig:comparison}\textbf{Comparison} of qualitative results of using Prompt Sliders against Concept Sliders \cite{gandikota2023sliders} for $\alpha=1$. Unlike Concept Sliders which require careful hyperparameter tuning for learning the target concept, Prompt Sliders are simple to optimize and work well for diverse prompts. The corresponding prompts for the images in the figure from left to right are as follows. \textit{"A whimsical dragon, full of texture and detail", "A bustling city street, with people walking", "A cat lounging comfortably on a sofa", "A child opening a gift", "A child playing with a favorite toy"}.}
\end{figure*}

\begin{figure*}[t]\RawFloats
\centering
\includegraphics[width=\columnwidth, trim= 0 160 0 140, clip]{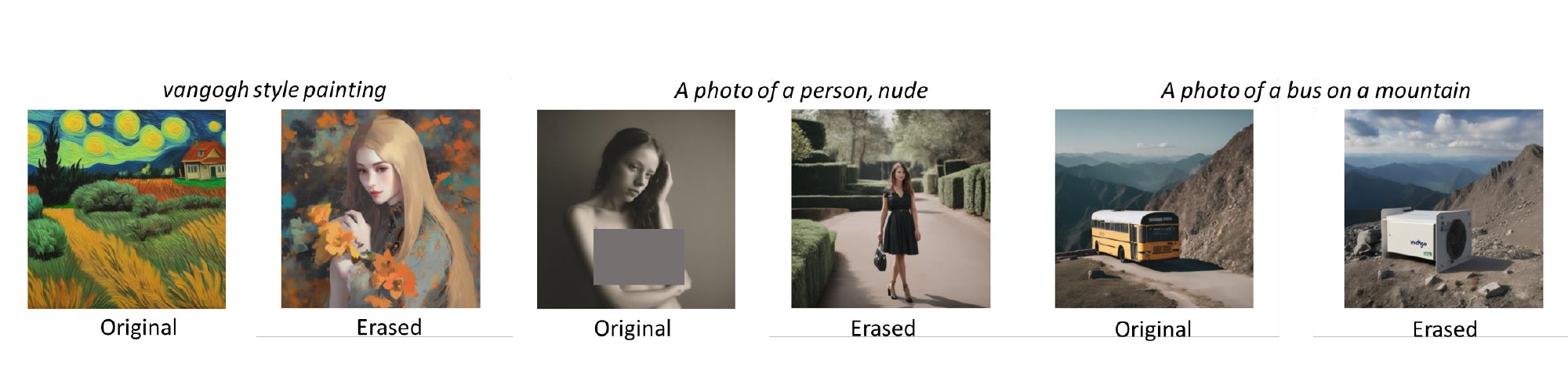}
\caption{\label{fig:erasing}\textbf{Erasing Concepts} from pretrained diffusion models using Prompt Sliders. The prompts used to generate the image are shown on top of each image.}
\end{figure*}

\subsection{\bf Comparison with Concept Sliders:}
Figure \ref{fig:comparison} shows the comparison of qualitative results between concept sliders and prompt sliders for the same prompt and seed. Concept sliders require careful hyperparameter tuning to effectively learn low-rank parameter directions for enhancing the target concept, and they may fail to capture the target concept otherwise, as illustrated in Figure \ref{fig:comparison}. In contrast, prompt sliders are easy to optimize and generalize much better across different prompts.

\subsection{\bf Erasing Concepts}
To erase concepts associated with a given target identifier, we learn the text embedding for that identifier rather than using a new token. This ensures that the concept is not generated, even if it is explicitly mentioned in the prompt.
Figure \ref{fig:erasing} shows that prompt sliders are effective in erasing concepts from the pre-trained model which is also generalizable to other models sharing the same text encoder. Figure \ref{fig:erasing} shows three such instances of erasing concepts such as style (vangogh), mature content (nudity) and object (bus).

\subsection{\bf Transferring to SD v1.4 and v1.5 models: }
\begin{figure*}[t]\RawFloats
\centering
\includegraphics[width=\columnwidth, trim= 0 120 0 120, clip]{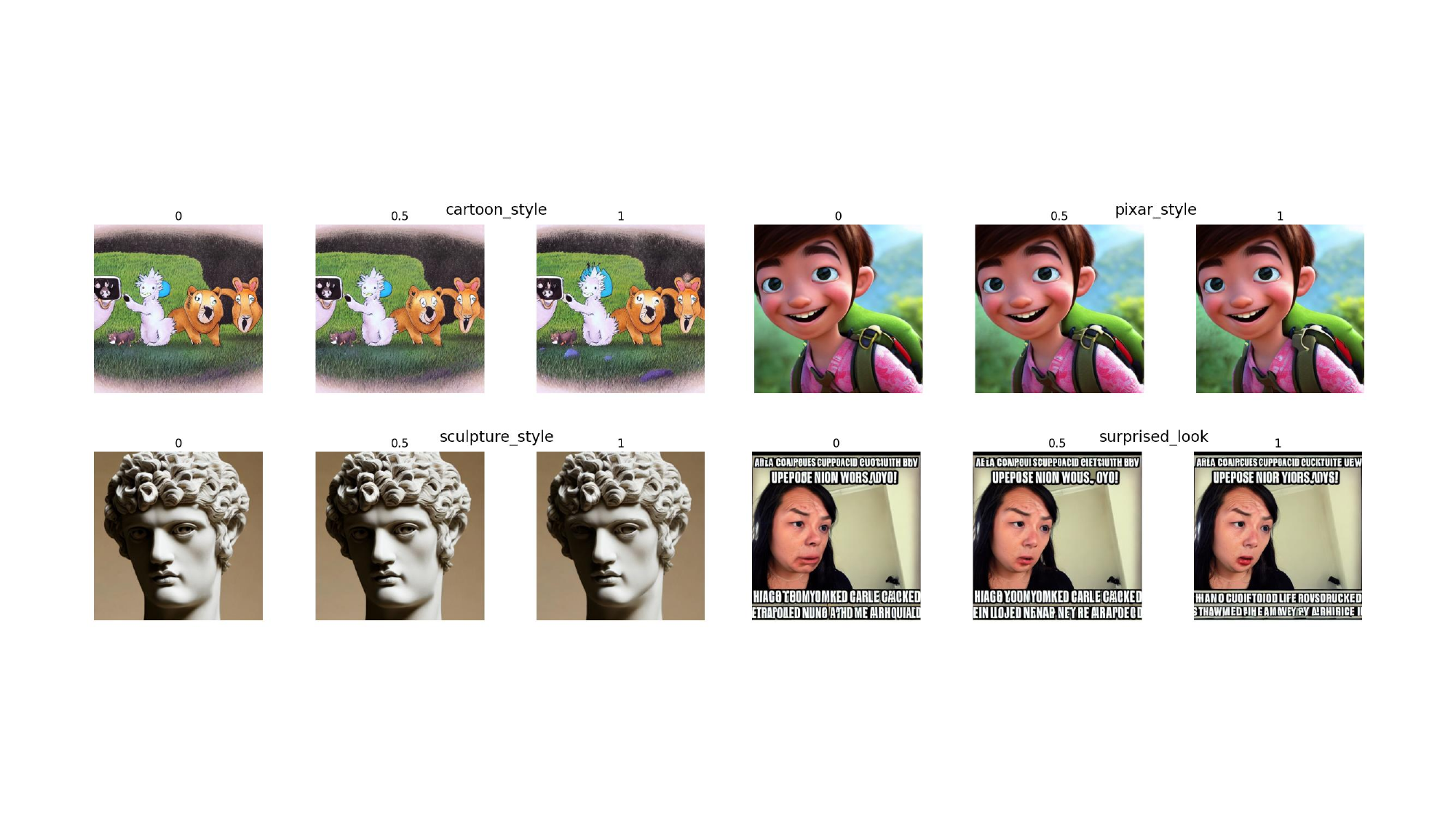}
\caption{\label{fig:sd14transfer}\textbf{Qualitative results} of transferring the trained prompt sliders from SD-XL to SD v1.4 model. The corresponding prompts for the images in the figure from left to right in the top row, and from left to right in the bottom row are as follows. \textit{"A family of animals having a fun adventure together", "A brave young hero embarking on an adventure, with expressive eyes and a big smile", "A Greek god with intricate details and lifelike expressions", "A person caught off-guard by unexpected news"}.}
\end{figure*}

\begin{figure*}[t]\RawFloats
\centering
\includegraphics[width=\columnwidth, trim= 0 120 0 120, clip]{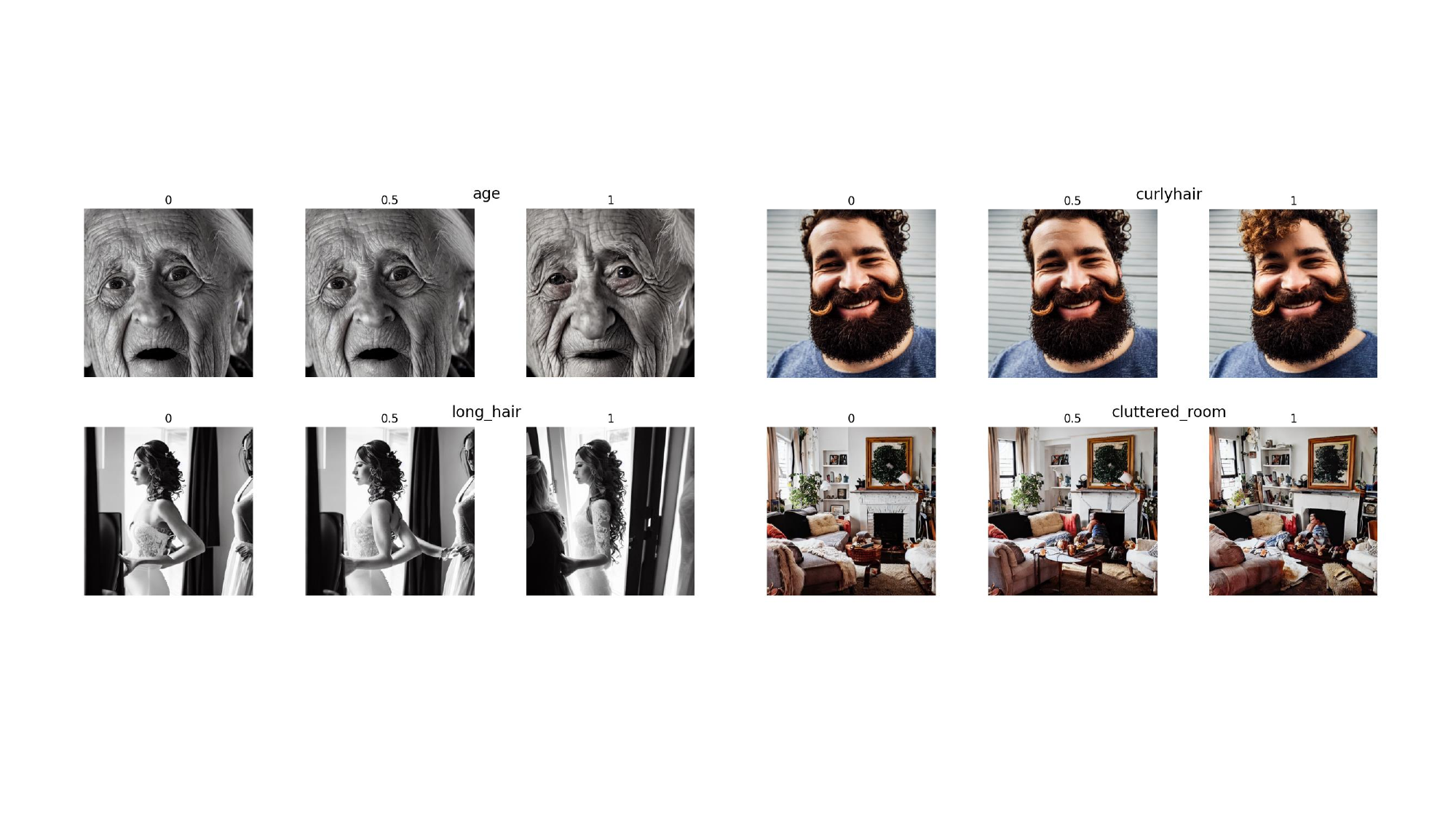}
\caption{\label{fig:sd15transfer}\textbf{Qualitative results} of transferring the trained prompt sliders from SD-XL to SD v1.5 model. The corresponding prompts for the images in the figure from left to right in the top row, and from left to right in the bottom row are as follows. \textit{"A photo of a person", "A man with a thick, beard, giving a charming smile", "A bride getting ready for her wedding day", "A cozy living room"}.}
\end{figure*}

Unlike prior methods \cite{gandikota2023sliders, gandikota2024unified} that require re-training of LoRA adapters for each diffusion model, Prompt Sliders generalize well to all models sharing the same text encoder.
Figures \ref{fig:sd14transfer} and \ref{fig:sd15transfer} illustrate the qualitative results of using the trained concept embeddings from SD-XL model to SD-v1.4 and SD-v1.5 models respectively. These concept embeddings capture specific directions in the textual space associated with the target concept, allowing them to generalize effectively across models that share the same text encoder. The bottom part of Table\ref{tab:speed} shows that the learned prompt sliders transferred from SD-XL model obtains comparable performance to the prompt sliders learned by Stable Diffusion v1.5 due to the shared clip text encoder used by both the models.

\begin{figure*}[t]\RawFloats
\centering
\includegraphics[width=\columnwidth, trim= 0 120 0 120, clip]{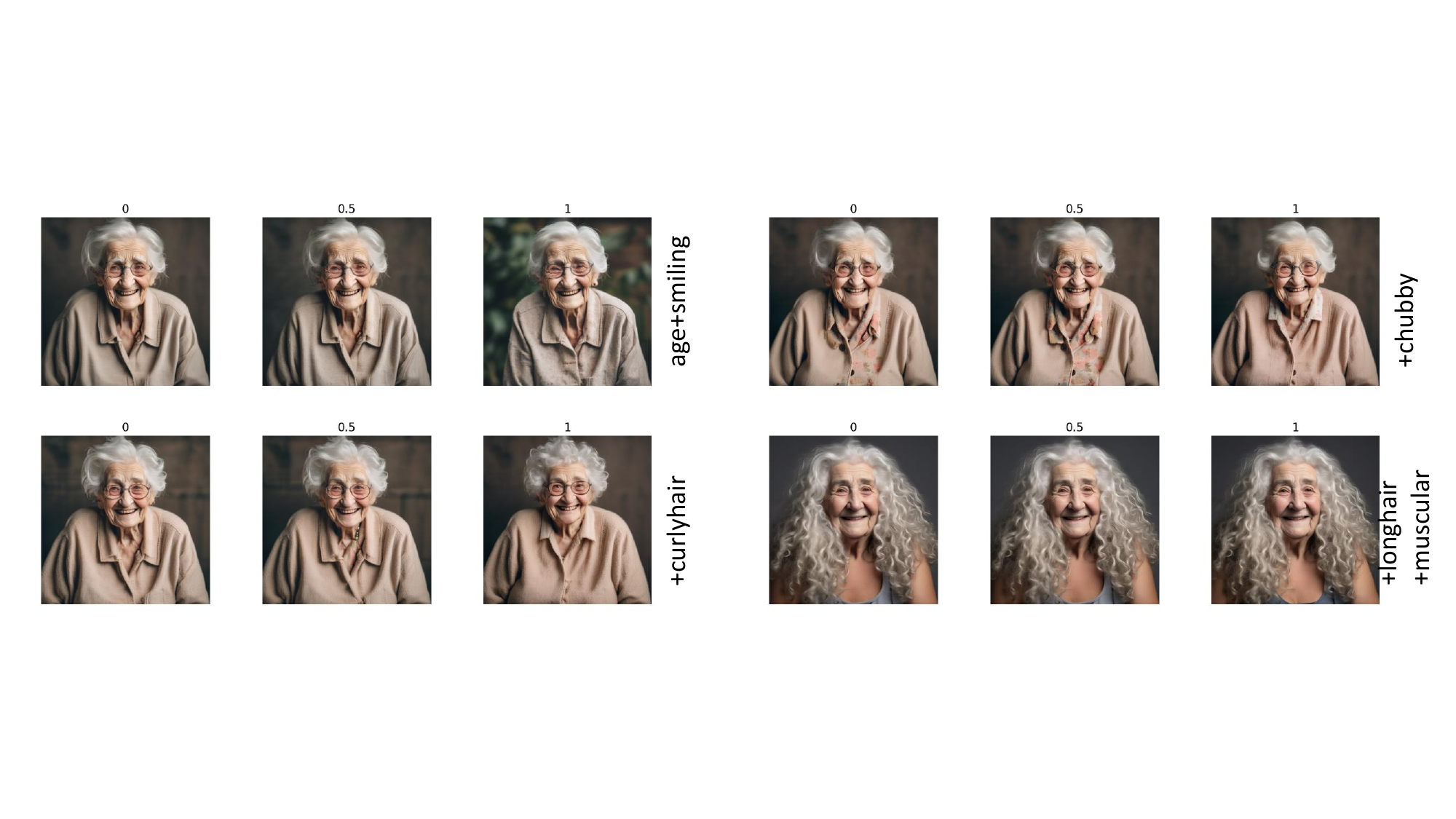}
\caption{\label{fig:multicompose}\textbf{Qualitative results} of composing multiple concepts together by appending them to the input prompt. The input prompt for all the images are "A photo of a person" followed by the concept tokens. Concepts are appended to the prompt sequentially and is depicted in the figure from left to right in the top row, then continues from left to right in the bottom row.}
\end{figure*}

\subsection{\bf Multi-Concept Composition}
Since the learned concept embeddings are in the textual space, it is straightforward to combine multiple concept embeddings by  appending it to the user input prompt. Figure \ref{fig:multicompose} shows the results of combining multiple concepts. New concepts are introduced sequentially from left to right in the top row, then continue from left to right in the bottom row. Figure \ref{fig:multicompose} also shows that the learned embeddings are well separated, even for closely related concepts such as curly hair and long hair, making it possible to generate attributes both independently  or together.

\begin{figure*}[!ht]\RawFloats
\centering
\includegraphics[width=\columnwidth, trim= 20 200 20 200, clip]{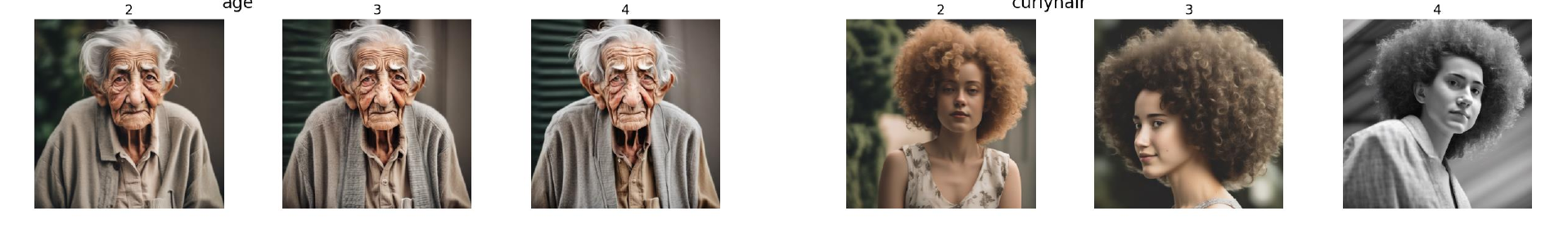}
\caption{\label{fig:limitation}\textbf{Qualitative results with increased guidance strength} Prompt sliders at higher guidance strength $\alpha$ overpower the overall generation, altering the image context while maintaining the subject identity. The effect is more noticeable for certain concepts like "curlyhair" while it is subtler for others such as "age".}
\end{figure*}

\subsection{Higher Guidance Strength}
At higher guidance strengths, the concepts learned by the prompt slider dominate the overall generation, altering the image context while maintaining the core identity, as demonstrated in Figure \ref{fig:limitation}. In this figure, it is evident that the effect is more pronounced for certain concepts like "curlyhair," where the pose of the subject changes significantly. Conversely, the effect is subtler for other concepts such as "age," where the background experiences only minor alterations. This indicates that while the prompt slider can effectively modify certain attributes, its influence varies depending on the specific concept being adjusted. Understanding these nuances is crucial for optimizing the use of guidance weights in achieving the desired balance between context modification and identity preservation, presenting an interesting direction for future research.

\section{Conclusion}
In this work, we introduced the Prompt Slider method for precise manipulation, editing, and erasure of concepts in diffusion models. Unlike previous approaches, prompt sliders do not introduce additional computational overhead or increase inference time, requiring only 3KB of storage space for each concept. Additionally, the learned concept embeddings are transferable across all models that use the same text encoder as the one used during training. Our results demonstrate that prompt sliders enable easy composition of various concepts, provide effective and precise control, editing, and erasure of concepts within diffusion models. This method not only enhances the flexibility and efficiency of concept manipulation but also offers a scalable solution for various applications in image generation. Future research could explore different textual inversion methods for learning the concepts and extending this technique to more complex scenarios.

\section*{Acknowledgements}
This work was partially funded by the NSF award IIS-2303153. We also acknowledge and thank the use of the Nautilus platform for some of the experiments discussed above.

{\small
\bibliographystyle{splncs04}
\bibliography{egbib}

\begin{thebibliography}{10}
\providecommand{\url}[1]{\texttt{#1}}
\providecommand{\urlprefix}{URL }
\providecommand{\doi}[1]{https://doi.org/#1}

\bibitem{Agustsson2017}
Agustsson, E., Mentzer, F., Tschannen, M., Cavigelli, L., Timofte, R., Gool, L.V.: Soft-to-hard vector quantization for end-to-end learning compressible representations. In: Advances in Neural Information Processing Systems (2017)

\bibitem{breakascene}
Avrahami, O., Aberman, K., Fried, O., Cohen-Or, D., Lischinski, D.: Break-a-scene: Extracting multiple concepts from a single image. In: SIGGRAPH Asia 2023 Conference Papers. SA '23, Association for Computing Machinery, New York, NY, USA (2023). \doi{10.1145/3610548.3618154}, \url{https://doi.org/10.1145/3610548.3618154}

\bibitem{brock2018large}
Brock, A., Donahue, J., Simonyan, K.: Large scale gan training for high fidelity natural image synthesis. arXiv preprint arXiv:1809.11096  (2018)

\bibitem{brooks2022instructpix2pix}
Brooks, T., Holynski, A., Efros, A.A.: Instructpix2pix: Learning to follow image editing instructions. In: CVPR (2023)

\bibitem{chen2023subjectdriven}
Chen, W., Hu, H., LI, Y., Ruiz, N., Jia, X., Chang, M.W., Cohen, W.W.: Subject-driven text-to-image generation via apprenticeship learning. In: Thirty-seventh Conference on Neural Information Processing Systems (2023), \url{https://openreview.net/forum?id=wv3bHyQbX7}

\bibitem{eccv2022_palavra_cohen}
Cohen, N., Gal, R., Meirom, E.A., Chechik, G., Atzmon, Y.: "this is my unicorn, fluffy": Personalizing frozen vision-language representations. In: European Conference on Computer Vision (ECCV) (2022)

\bibitem{dhariwal2021diffusion}
Dhariwal, P., Nichol, A.: Diffusion models beat gans on image synthesis. Advances in Neural Information Processing Systems  \textbf{34},  8780--8794 (2021)

\bibitem{gal2022textual}
Gal, R., Alaluf, Y., Atzmon, Y., Patashnik, O., Bermano, A.H., Chechik, G., Cohen-Or, D.: An image is worth one word: Personalizing text-to-image generation using textual inversion. arXiv  (2022). \doi{10.48550/ARXIV.2208.01618}, \url{https://arxiv.org/abs/2208.01618}

\bibitem{gal2023encoderbaseddomaintuningfast}
Gal, R., Arar, M., Atzmon, Y., Bermano, A.H., Chechik, G., Cohen-Or, D.: Encoder-based domain tuning for fast personalization of text-to-image models (2023), \url{https://arxiv.org/abs/2302.12228}

\bibitem{gandikota2023erasing}
Gandikota, R., Materzy\'nska, J., Fiotto-Kaufman, J., Bau, D.: Erasing concepts from diffusion models. In: Proceedings of the 2023 IEEE International Conference on Computer Vision (2023)

\bibitem{gandikota2023sliders}
Gandikota, R., Materzy\'nska, J., Zhou, T., Torralba, A., Bau, D.: Concept sliders: Lora adaptors for precise control in diffusion models. arXiv preprint arXiv:2311.12092  (2023)

\bibitem{gandikota2024unified}
Gandikota, R., Orgad, H., Belinkov, Y., Materzy\'nska, J., Bau, D.: Unified concept editing in diffusion models. IEEE/CVF Winter Conference on Applications of Computer Vision  (2024)

\bibitem{Geng23instructdiff}
Geng, Z., Yang, B., Hang, T., Li, C., Gu, S., Zhang, T., Bao, J., Zhang, Z., Hu, H., Chen, D., Guo, B.: Instructdiffusion: {A} generalist modeling interface for vision tasks. CoRR  \textbf{abs/2309.03895} (2023). \doi{10.48550/arXiv.2309.03895}, \url{https://doi.org/10.48550/arXiv.2309.03895}

\bibitem{goodfellow_2014_gan}
Goodfellow, I., Pouget-Abadie, J., Mirza, M., Xu, B., Warde-Farley, D., Ozair, S., Courville, A., Bengio, Y.: Generative adversarial nets. In: Ghahramani, Z., Welling, M., Cortes, C., Lawrence, N., Weinberger, K. (eds.) Advances in Neural Information Processing Systems. vol.~27. Curran Associates, Inc. (2014), \url{https://proceedings.neurips.cc/paper_files/paper/2014/file/5ca3e9b122f61f8f06494c97b1afccf3-Paper.pdf}

\bibitem{han2023svdiffcompactparameterspace}
Han, L., Li, Y., Zhang, H., Milanfar, P., Metaxas, D., Yang, F.: Svdiff: Compact parameter space for diffusion fine-tuning (2023), \url{https://arxiv.org/abs/2303.11305}

\bibitem{hertz2022prompt}
Hertz, A., Mokady, R., Tenenbaum, J., Aberman, K., Pritch, Y., Cohen-Or, D.: Prompt-to-prompt image editing with cross attention control. arXiv preprint arXiv:2208.01626  (2022)

\bibitem{ddpm_neurips_20}
Ho, J., Jain, A., Abbeel, P.: Denoising diffusion probabilistic models. In: Larochelle, H., Ranzato, M., Hadsell, R., Balcan, M., Lin, H. (eds.) Advances in Neural Information Processing Systems. vol.~33, pp. 6840--6851. Curran Associates, Inc. (2020), \url{https://proceedings.neurips.cc/paper_files/paper/2020/file/4c5bcfec8584af0d967f1ab10179ca4b-Paper.pdf}

\bibitem{ho2020denoising}
Ho, J., Jain, A., Abbeel, P.: Denoising diffusion probabilistic models. Advances in Neural Information Processing Systems  \textbf{33},  6840--6851 (2020)

\bibitem{hu2022lora}
Hu, E.J., yelong shen, Wallis, P., Allen-Zhu, Z., Li, Y., Wang, S., Wang, L., Chen, W.: Lo{RA}: Low-rank adaptation of large language models. In: International Conference on Learning Representations (2022), \url{https://openreview.net/forum?id=nZeVKeeFYf9}

\bibitem{sdxllorainference}
HuggingFace: https://huggingface.co/blog/lora-adapters-dynamic-loading (2023), webpage

\bibitem{sdxltricks2023}
Inui, N.: Sd/sdxl tricks beneath the papers and codes (2023), github

\bibitem{jia2023tamingencoderzerofinetuning}
Jia, X., Zhao, Y., Chan, K.C.K., Li, Y., Zhang, H., Gong, B., Hou, T., Wang, H., Su, Y.C.: Taming encoder for zero fine-tuning image customization with text-to-image diffusion models (2023), \url{https://arxiv.org/abs/2304.02642}

\bibitem{karras2019style}
Karras, T., Laine, S., Aila, T.: A style-based generator architecture for generative adversarial networks. In: Proceedings of the IEEE/CVF conference on computer vision and pattern recognition. pp. 4401--4410 (2019)

\bibitem{Karras_2020_CVPR}
Karras, T., Laine, S., Aittala, M., Hellsten, J., Lehtinen, J., Aila, T.: Analyzing and improving the image quality of stylegan. In: Proceedings of the IEEE/CVF Conference on Computer Vision and Pattern Recognition (CVPR) (June 2020)

\bibitem{kawar2023imagic}
Kawar, B., Zada, S., Lang, O., Tov, O., Chang, H., Dekel, T., Mosseri, I., Irani, M.: Imagic: Text-based real image editing with diffusion models. In: Conference on Computer Vision and Pattern Recognition 2023 (2023)

\bibitem{Kim_2022_CVPR}
Kim, G., Kwon, T., Ye, J.C.: Diffusionclip: Text-guided diffusion models for robust image manipulation. In: Proceedings of the IEEE/CVF Conference on Computer Vision and Pattern Recognition (CVPR). pp. 2426--2435 (June 2022)

\bibitem{kim2023safeselfdistillationinternetscaletexttoimage}
Kim, S., Jung, S., Kim, B., Choi, M., Shin, J., Lee, J.: Towards safe self-distillation of internet-scale text-to-image diffusion models (2023), \url{https://arxiv.org/abs/2307.05977}

\bibitem{kingma2013auto}
Kingma, D.P., Welling, M.: {Auto-Encoding Variational Bayes}. In: 2nd International Conference on Learning Representations, {ICLR} 2014, Banff, AB, Canada, April 14-16, 2014, Conference Track Proceedings (2014)

\bibitem{kumari2022customdiffusion}
Kumari, N., Zhang, B., Zhang, R., Shechtman, E., Zhu, J.Y.: Multi-concept customization of text-to-image diffusion. In: Proceedings of the IEEE/CVF Conference on Computer Vision and Pattern Recognition (CVPR) (2023)

\bibitem{meng2022sdedit}
Meng, C., He, Y., Song, Y., Song, J., Wu, J., Zhu, J.Y., Ermon, S.: {SDE}dit: Guided image synthesis and editing with stochastic differential equations. In: International Conference on Learning Representations (2022)

\bibitem{nichol2022glidephotorealisticimagegeneration}
Nichol, A., Dhariwal, P., Ramesh, A., Shyam, P., Mishkin, P., McGrew, B., Sutskever, I., Chen, M.: Glide: Towards photorealistic image generation and editing with text-guided diffusion models (2022), \url{https://arxiv.org/abs/2112.10741}

\bibitem{nichol2021improved}
Nichol, A.Q., Dhariwal, P.: Improved denoising diffusion probabilistic models. In: International Conference on Machine Learning. pp. 8162--8171. PMLR (2021)

\bibitem{nitzan2022mystyle}
Nitzan, Y., Aberman, K., He, Q., Liba, O., Yarom, M., Gandelsman, Y., Mosseri, I., Pritch, Y., Cohen-Or, D.: Mystyle: A personalized generative prior. arXiv preprint arXiv:2203.17272  (2022)

\bibitem{VanDenOord2017}
van~den Oord, A., Vinyals, O., Kavukcuoglu, K.: Neural discrete representation learning. In: Advances in Neural Information Processing Systems (2017)

\bibitem{park2023understanding}
Park, Y.H., Kwon, M., Choi, J., Jo, J., Uh, Y.: Understanding the latent space of diffusion models through the lens of riemannian geometry. In: Thirty-seventh Conference on Neural Information Processing Systems (2023), \url{https://openreview.net/forum?id=VUlYp3jiEI}

\bibitem{zero-shot-im-im-parmar}
Parmar, G., Kumar~Singh, K., Zhang, R., Li, Y., Lu, J., Zhu, J.Y.: Zero-shot image-to-image translation. In: ACM SIGGRAPH 2023 Conference Proceedings. SIGGRAPH '23, Association for Computing Machinery, New York, NY, USA (2023). \doi{10.1145/3588432.3591513}, \url{https://doi.org/10.1145/3588432.3591513}

\bibitem{podell2023sdxlimprovinglatentdiffusion}
Podell, D., English, Z., Lacey, K., Blattmann, A., Dockhorn, T., Müller, J., Penna, J., Rombach, R.: Sdxl: Improving latent diffusion models for high-resolution image synthesis (2023), \url{https://arxiv.org/abs/2307.01952}

\bibitem{radford2021learning}
Radford, A., Kim, J., Hallacy, C., Ramesh, A., Goh, G., Agarwal, S., Sastry, G., Askell, A., Mishkin, P., Clark, J., et~al.: Learning transferable visual models from natural language supervision. In: International conference on machine learning. pp. 8748--8763 (2021)

\bibitem{Ramesh2022HierarchicalTI}
Ramesh, A., Dhariwal, P., Nichol, A., Chu, C., Chen, M.: Hierarchical text-conditional image generation with clip latents. ArXiv  \textbf{abs/2204.06125} (2022), \url{https://api.semanticscholar.org/CorpusID:248097655}

\bibitem{vae-rezende14}
Rezende, D.J., Mohamed, S., Wierstra, D.: Stochastic backpropagation and approximate inference in deep generative models. In: Xing, E.P., Jebara, T. (eds.) Proceedings of the 31st International Conference on Machine Learning. Proceedings of Machine Learning Research, vol.~32, pp. 1278--1286. PMLR, Bejing, China (22--24 Jun 2014)

\bibitem{rombach2022high}
Rombach, R., Blattmann, A., Lorenz, D., Esser, P., Ommer, B.: High-resolution image synthesis with latent diffusion models. In: Proceedings of the IEEE/CVF Conference on Computer Vision and Pattern Recognition. pp. 10684--10695 (2022)

\bibitem{ronneberger2015u}
Ronneberger, O., Fischer, P., Brox, T.: U-net: Convolutional networks for biomedical image segmentation. In: Medical Image Computing and Computer-Assisted Intervention--MICCAI 2015: 18th International Conference, Munich, Germany, October 5-9, 2015, Proceedings, Part III 18. pp. 234--241. Springer (2015)

\bibitem{ruiz2022dreambooth}
Ruiz, N., Li, Y., Jampani, V., Pritch, Y., Rubinstein, M., Aberman, K.: Dreambooth: Fine tuning text-to-image diffusion models for subject-driven generation. CVPR  (2023)

\bibitem{ryu2022}
Ryu, S.: https://github.com/cloneofsimo/lora (2022), github

\bibitem{Shi_2024_CVPR}
Shi, J., Xiong, W., Lin, Z., Jung, H.J.: Instantbooth: Personalized text-to-image generation without test-time finetuning. In: Proceedings of the IEEE/CVF Conference on Computer Vision and Pattern Recognition (CVPR). pp. 8543--8552 (June 2024)

\bibitem{shi2023dragdiffusion}
Shi, Y., Xue, C., Pan, J., Zhang, W., Tan, V.Y., Bai, S.: Dragdiffusion: Harnessing diffusion models for interactive point-based image editing. arXiv preprint arXiv:2306.14435  (2023)

\bibitem{nethermo-sohl-dickstein15}
Sohl-Dickstein, J., Weiss, E., Maheswaranathan, N., Ganguli, S.: Deep unsupervised learning using nonequilibrium thermodynamics. In: Bach, F., Blei, D. (eds.) Proceedings of the 32nd International Conference on Machine Learning. Proceedings of Machine Learning Research, vol.~37, pp. 2256--2265. PMLR, Lille, France (07--09 Jul 2015)

\bibitem{tewel2023keylocked}
Tewel, Y., Gal, R., Chechik, G., Atzmon, Y.: Key-locked rank one editing for text-to-image personalization. In: ACM SIGGRAPH 2023 Conference Proceedings. SIGGRAPH '23 (2023)

\bibitem{vaswani2017attention}
Vaswani, A., Shazeer, N., Parmar, N., Uszkoreit, J., Jones, L., Gomez, A.N., Kaiser, {\L}., Polosukhin, I.: Attention is all you need. Advances in neural information processing systems  \textbf{30} (2017)

\bibitem{voynov2023P+}
Voynov, A., Chu, Q., Cohen-Or, D., Aberman, K.: P+: Extended textual conditioning in text-to-image generation. In: arXiv preprint (2023)

\bibitem{fingers}
Walker, J.: {AI Explain Why They're So Bad At Drawing Human Fingers}. \url{https://kotaku.com/chatgpt-chatsonic-ai-dall-e-render-human-fingers-why-1850107682} (2023)

\bibitem{wei2023elite}
Wei, Y., Zhang, Y., Ji, Z., Bai, J., Zhang, L., Zuo, W.: Elite: Encoding visual concepts into textual embeddings for customized text-to-image generation. arXiv preprint arXiv:2302.13848  (2023)

\bibitem{xu2022poisson}
Xu, Y., Liu, Z., Tegmark, M., Jaakkola, T.S.: Poisson flow generative models. In: Oh, A.H., Agarwal, A., Belgrave, D., Cho, K. (eds.) Advances in Neural Information Processing Systems (2022), \url{https://openreview.net/forum?id=voV_TRqcWh}

\bibitem{pmlr-v202-xu23m}
Xu, Y., Liu, Z., Tian, Y., Tong, S., Tegmark, M., Jaakkola, T.: {PFGM}++: Unlocking the potential of physics-inspired generative models. In: Proceedings of the 40th International Conference on Machine Learning. pp. 38566--38591 (2023)

\bibitem{yadav2023tiesmerging}
Yadav, P., Tam, D., Choshen, L., Raffel, C., Bansal, M.: {TIES}-merging: Resolving interference when merging models. In: Thirty-seventh Conference on Neural Information Processing Systems (2023), \url{https://openreview.net/forum?id=xtaX3WyCj1}

\bibitem{yu2024languagedare}
Yu, L., Yu, B., Yu, H., Huang, F., Li, Y.: Language models are super mario: Absorbing abilities from homologous models as a free lunch. In: International Conference on Machine Learning. PMLR (2024)

\bibitem{leco2023}
Zhou, T.: https://github.com/p1atdev/leco (2023), github

\end{thebibliography}
}
\end{document}